\documentclass[11pt,a4paper]{article}
\usepackage[hyperref]{acl2018}
\usepackage{times}
\usepackage{latexsym}
\usepackage{amsmath}
\usepackage{url}
\usepackage{array, caption, tabularx, makecell, booktabs}%
\usepackage{multirow}
\usepackage{color}
\usepackage{colortbl}
\usepackage[colorlinks]{}
\usepackage{subcaption}
\usepackage{graphicx}

\aclfinalcopy 

\title{Neural Generation of Diverse Questions \\ using Answer Focus, Contextual and Linguistic Features 
}

\author{Vrindavan Harrison \textnormal{and} Marilyn Walker \\
  University of California Santa Cruz \\
  Santa Cruz, CA, US \\
  {\tt \{vharriso, mawalker\}@ucsc.edu }\\ }

\date{}

\begin{document}
\maketitle

\begin{abstract}
Question Generation is the task of automatically creating questions
from textual input. In this work we present a new Attentional
Encoder--Decoder Recurrent Neural Network model for automatic question
generation. Our model incorporates linguistic
features and an additional sentence embedding to capture meaning at both sentence and word levels. The linguistic features are designed to capture information related to named entity recognition, word case, and entity coreference resolution. 
In addition our model uses a copying mechanism and a special answer signal that enables generation of numerous diverse questions on a given sentence. 
Our model achieves state of the art results of 19.98 Bleu\_4 on a benchmark Question Generation dataset, outperforming all previously published results by a
significant margin. A human evaluation also shows that the added features improve the quality of the generated questions.
\end{abstract}

\section{Introduction}

Question Generation (QG) is the task of automatically generating questions from textual input \cite{Rus_Wyse_Piwek_Lintean_Stoyanchev_Moldovan_2010}. There are a wide variety of question types and forms,  e.g., short answer, open ended, multiple choice, and gap questions, each require a different approach to generate. One distinguishing aspect of a QG system is the type of questions that it produces. This paper focuses on the generation of factoid short answer questions, i.e., questions that can be answered by a single short phrase, usually appearing directly in the input text. 

The work of a QG system typically consists of three conceptual
subtasks: Target Selection, Question Representation Construction, and
Question Realization.
In Target Selection, important sentences and words within those sentences are identified.
During Question Representation Construction, suitable question--type and syntactic form are determined based on the characteristics of the sentence at hand and the words it contains. An example of this can be seen in \citet{Agarwal_Shah_Mannem_2011} who define rules based on the discourse connectives in a sentence to decide which question--type is most appropriate.
In the Question Realization step, the final surface form of the question is created.

It is common for QG systems to use a combination of semantic pattern matching, syntactic features, and template methods to create questions. Typically these systems look for patterns of syntax, keywords, or semantic roles that appear in the input sentence. Then they use these patterns to choose an appropriate question template, or use syntactic features to perform manipulations on the sentence to produce a question. 

These rule-based systems have some strengths over Neural Network models: they are easier to interpret and allow developers greater control over model behavior. Furthermore, they typically require less data to develop than a complex Neural Network might need to achieve a similar level of performance. 
However, rule-based systems have some weaknesses as well. They tend to be laborious to develop, or domain specific. For example the system developed by \citet{MOSTOW_CHEN_2009} relies on the presence of one of a set of 239 modal verbs in a sentence, and \citet{Olney_Graesser_Person_2012} use 3000 keywords provided by the glossary of a Biology text book and a test-prep study guide. The system described in \citet{Chali_Hasan_2015} uses roughly 350 hand-crafted rules. Furthermore, these systems rely heavily on syntactic parsers, and may struggle to recover from parser inaccuracies.

Among many different approaches to question generation, our
work is most similar to recent work applying neural network models to
the task of generating short answer factoid questions for SQUAD
\cite{Du_Shao_Cardie_2017, Yuan_Wang_Gulcehre_Sordoni_Bachman_Zhang_Subramanian_Trischler_2017, Sachan_Xing_2018}. However these previous models have several limitations. As illustrated
in Table~\ref{table:sentence-question-examples-1}, the SQUAD corpus \cite{Rajpurkar_Zhang_Lopyrev_Liang_2016}
provides multiple gold standard references for each sentence (Q1, Q2,
and Q3), but previous work to date can only generate one question for
each sentence as represented by the baseline model (Q4), whereas our model
can generate multiple questions as shown in Table \ref{table:sentence-question-examples-1}.

\begin{table}[t]
\begin{small}
\begin{tabular}
{@{} p{0.1in}|p{2.8in}@{}}
\toprule
\rowcolor[gray]{0.9} {\bf \#} & 
\multicolumn{1}{c}{ {\bf Sentence}}  \\
S1 & The character of midna has the most voice acting -- her on-screen dialog is often accompanied by a babble of pseudo-speech , which was produced by scrambling the phonemes of english phrases [ better source needed ] sampled by japanese voice actress akiko komoto. \\ \hline

\rowcolor[gray]{0.9}\multicolumn{2}{c}{{\bf Gold Standard}}   \\  
Q1 &which person has the most spoken dialogue in the game? \\ \hline   
Q2 & who provided the basis for midna's voice? \\ \hline   
Q3 & what country does akiko komoto come from? \\ \hline   

\rowcolor[gray]{0.9}\multicolumn{2}{c}{ {\bf Baseline}} \\  
Q4 & what is her ? \\ \hline   
\rowcolor[gray]{0.9}\multicolumn{2}{c}{ {\bf Our Model: FocusCR}}  \\ \hline      
Q5 & what character has the most voice acting in english? \\ \hline   
Q6 &  what is the name of the japanese voice actress?\\ \hline   
Q7 & what is the nationality of akiko komoto? \\ \hline   
\end{tabular}
\end{small}
\vspace{-.1in}
\caption{Sentence and associated questions generated from the baseline and our best model.\label{table:sentence-question-examples-1}}
\end{table}

In Section~\ref{model-sec}, we present our novel model that introduces
additional token supervision representing features of the text as well
as an additional lower dimensional word embedding. The features
include a Named Entity Recognition (NER) feature, a word case feature,
and a special answer signaling feature.  The answer signaling feature
allows our model to generate multiple questions for each sentence,
illustrated with Q5, Q6 and Q7 in
Table~\ref{table:sentence-question-examples-1}.  We also introduce a
coreference resolution model and supplement the sentence input
representation with resolved coreferences, as well as a copying
mechanism. Section~\ref{results-sec} presents an evaluation of the
final model on the benchmark SQuAD testset using automatic evaluation
metrics and  shows that it achieves state of the art results of 19.98
BLEU\_4, 22.26 METEOR, and 48.23 ROUGE 
\cite{papineni2002bleu,banerjee2005meteor,lin2004rouge}. To our
knowledge this model outperforms all previously published results by a
significant margin. A human evaluation also shows that the introduced
features and answer-specific sentence embedding
improve the quality of the
generated questions. We delay a more detailed review of previous work
to Section~\ref{rel-sec} and conclude in Section~\ref{conc-sec}.

\section{Model}
\label{model-sec}

Our QG model follows a standard RNN Encoder--Decoder model
\cite{sutskever2014sequence} that maps a source sequence 
(declarative sentence) to a target sequence (question).  The
architecture of the baseline model is as follows: the encoder is a
multi-layer bidirectional LSTM \cite{hochreiter1997long} and the
decoder is a uni-directional LSTM that uses global attention with
input-feeding \cite{luong2015effective}. This baseline model yields one question per sentence (Q4 in
Table~\ref{table:sentence-question-examples-1}).

\begin{figure}[t!h]
    \includegraphics[width=\linewidth]{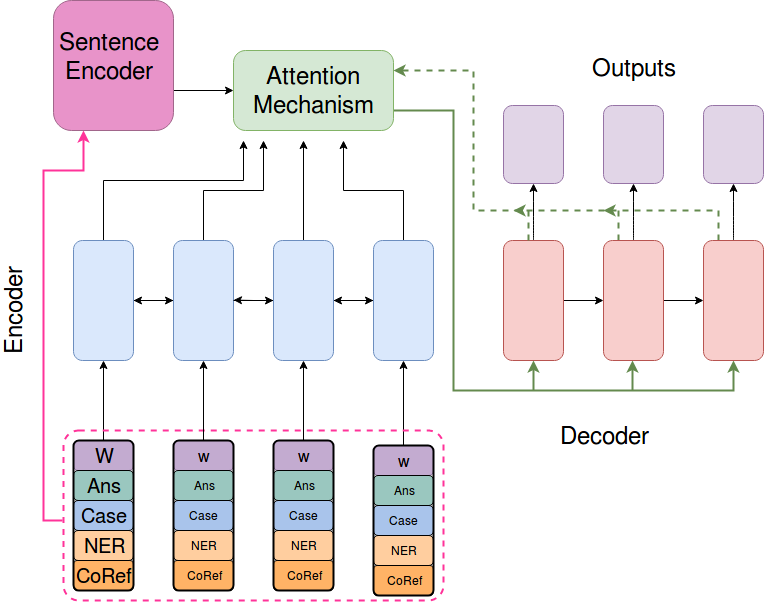}
    \caption{Diagram of our answer focus model.}
    \label{fig:model_diagram}
\end{figure}

 We create our model by enhancing the baseline model in the following three ways: 
\begin{itemize}
\item We add 4 different token level supervision features to the input. See Section~\ref{feat-sec}.
\item We add a sentence encoder that creates a question specific sentence embedding. 
\item We use a copy mechanism \cite{see2017get} to copy words from the sentence directly into the question.
\end{itemize}

\subsection{Feature Supervision}
\label{feat-sec}

A feature-rich encoding is constructed by concatenating several token level features onto the token's word-based embedding using a method similar to that described by \citet{Nallapati_Zhou_santos_Gulcehre_Xiang_2016} for abstractive text summarization. 

\begin{table}[h!t]
\begin{small}
\begin{tabular}
{@{} p{0.05in}|p{2.8in}@{}} \toprule \rowcolor[gray]{0.9} {\bf \#} &
\multicolumn{1}{c}{ {\bf Sentence w/ Feature Additions}} \\ S1 & the
character of {\bf midna}$_{a1}$({\sc ne = location}) has the most
voice acting -- her ({\sc coref = the character of midna}({\sc ne = location}))
on-screen dialog is often accompanied by a babble of pseudo-speech,
which was produced by scrambling the phonemes of english({\sc ne =
  nationality}) phrases [better source needed] sampled by {\bf
  japanese}$_{a3}$({\sc ne = nationality}) voice {\bf actress$_{a2}$}({\sc ne =title}) {\bf
  akiko$_{a2}$}({\sc ne = person}) {\bf komoto}$_{a2}$({\sc ne = person}) . \\ \hline

\end{tabular}
\end{small}
\vspace{-.1in}
\caption{Feature Markup on S1 \label{table:feature-examples-S1}}
\vspace{-.1in}
\end{table}

\noindent \textbf{Answer Signal Feature.} It is usually the case that  multiple questions can be asked about information contained within a single sentence. Therefore, a model that is capable of generating multiple questions for a  single sentence has greater utility than a model  such as the one described by \citet{Du_Shao_Cardie_2017} which is capable of generating only a single question per unique input sentence. This need to generate multiple questions for a sentence motivates our use of an answer signal. The model described by \citet{Yuan_Wang_Gulcehre_Sordoni_Bachman_Zhang_Subramanian_Trischler_2017} also uses an answer signal feature. However, by combining it with additional features and the question specific  sentence encoder  our model achieves better results, as we show in Section~\ref{results-sec}.

The answer signal is equivalent to the output of the target selection module in a standard QG pipeline; this is provided as part of the SQUAD corpus, but is straightforward to calculate automatically. 

The Answer Signal feature guides the model in deciding which information to focus on when reading the sentence. The signal being active in some location of the sentence indicates the answer to the question being generated. Then, modifying the location of the answer signal and keeping the rest of the sentence fixed enables the model to generate multiple answer specific questions per a given sentence.

The Answer Signal feature is implemented as a binary feature indicating whether or not a given token is part of the answer span. Table~\ref{table:feature-examples-S1} illustrates the results of the answer signal feature on S1 from Table~\ref{table:sentence-question-examples-1}. The answer signals are shown in {\bf bold} and annotated with an index. The indices $a1$, $a2$ and $a3$ correspond to Q5, Q6 and Q7, respectfully, from Table~\ref{table:sentence-question-examples-1} (as generated by our best model). 
For the sake of brevity we have simultaneously accentuated three separate answer signals in Table~\ref{table:feature-examples-S1}. In actuality, the model sees only one answer signal series per sentence input. To generate questions Q5, Q6, and Q7 the model was fed the same sentence three separate times, each time with only one of $a1$, $a2$ or $a3$ activated.

\noindent
\textbf{Case Feature.} The case feature is a simple binary feature that represents whether or not the input token contains an upper case letter. 

\noindent
\textbf{NER Feature.} We use an NER feature that is designed in the same fashion as \citet{Nallapati_Zhou_santos_Gulcehre_Xiang_2016}, who have previously used an NER based feature embedding to improve  performance of their sequence-to-sequence model used for Abstractive  Text Summarization. Just as in Abstractive Text Summarization, identifying important entities that are central to the meaning of a sentence is an imperative component of the QG task. 

The NER labels are computed in a  pre-processing step. The result of NER labeling performed on  a sentence is shown in Table~\ref{table:feature-examples-S1}. Similar to traditional word embeddings, we build a look-up  based embedding for each NER label. During execution, the  embedding associated with each token's NER label is retrieved using a table look-up and then concatenated onto the word embedding. The NER embeddings are a trainable parameter  which get updated during the model's training process. Figure~\ref{fig:model_diagram} shows a diagram depicting how the NER feature is incorporated into each token's feature  rich encoding via concatenation.

\noindent
\textbf{Coreference Feature.} Coreference labels are computed automatically in a pre-processing step. The coreference labels are calculated using all of the prior context for the input sentence text, but the input to the model is just the sentence augmented with the additional feature input. Table~\ref{table:feature-examples-S1} shows how the NER and coreference features are expectedly noisy. Nevertheless, they improve the model as we show below.

\begin{table}[h!t]
\begin{small}
\begin{tabular}
{@{} p{0.6in}|p{2.2in}@{}}
\toprule
\rowcolor[gray]{0.9} 
\multicolumn{2}{l} {{\bf \#} \hspace{2cm} {\bf Partial Context}}  \\
C1-33 &
In June 2014, Beyonc\'e ranked at \#1 on the Forbes Celebrity 100 list, earning an estimated \$115 million throughout June 2013 – June 2014. 
This in turn was the first time she had topped the Celebrity 100 list as well as being her highest yearly earnings to date.  \\ \hline

\rowcolor[gray]{0.9} \multicolumn{2}{c}{ {\bf Sentence}}  \\
S2 & As of May 2015, her net worth is estimated to be 
\$250 million. \\ \hline

\rowcolor[gray]{0.9}\multicolumn{2}{c}{{\bf Model Input with Coreference}}   \\  
S2 w/ coref& As of may({\sc ne = date}) 2015({\sc ne = date}), her ({\sc coref = beyonc\'e ({\sc ne = person})'s}) net worth is estimated to be \$({\sc ne = money}) 250({\sc ne = money}) million({\sc ne = money}). \\
Q8 & what is beyonc\'e's net worth in 2015? \\ \hline
\rowcolor[gray]{0.9}\multicolumn{2}{c}{{\bf Model Input w/out Coreference}}   \\  
S2 no coref & As of may({\sc ne = date}) 2015({\sc ne = date}), her net worth is estimated to be \$({\sc ne = money}) 250({\sc ne = money}) million({\sc ne = money}). \\
Q9 & what is elizabeth's net worth in may 2015 ?\\ \hline   
\end{tabular}
\end{small}
\vspace{-.1in}
\caption{Context, Sentence and Questions generated with and without
coreference.\label{table:coref-examples}}
\end{table}

Table~\ref{table:coref-examples} provides a detailed example 
for coreference showing context and the input sentence representation
with and without the coreference feature, as well as the effect on
the questions generated by the model. It is easy to see the benefits
of the coreference representation qualititively on individual examples.
For example, without coreference the model finds an entity {\it elizabeth}
who is associated with {\it net worth} in the language model and uses
that entity to generate the question rather than using {\it beyonc\'e},
the entity in context. In Section~\ref{results-sec} we show
that this qualitative difference affects the quantitative performance
measures. 

Coreference information is incorporated into the model by augmenting the 
input text as shown in Table~\ref{table:coref-examples}.
The representative mention of each entity gets inserted into the sentence 
following its coreferent. This results in two phrases referencing 
the same entity appearing in the text one immediately following the other, the first
being the one that appeared in the original text and the second
being the entity's representative mention. Each token is assigned 
a binary feature indicating whether the token was in the original text 
or if it has been thus inserted. This way the model can learn
to include or ignore the augmenting text as it deems necessary.

\subsection{Answer Focused Sentence Embedding}
\label{embed-sec}

The input sentence is encoded using a multi-layer bidirectional LSTM distinct from the token level encoder as illustrated in Figure~\ref{fig:model_diagram}. After completing the calculations of the last time step, the final state of the LSTM is taken as a sentence embedding. Then, this sentence embedding is concatenated on to the token level encoding for each time step. This allows the answer-specific sentence embedding to influence decoding decisions during each time step of the decoding process. 


We experiment with pre-training the sentence encoder. The pre-training process 
is carried out in two steps. First, to facilitate training of the sentence encoder
by itself, we need some ground truth sentence representation from which to measure 
similarity. Since this sentence encoder is used in question generation, it would 
be helpful to encode the sentence in such a way as to maximize its benefit to the 
QG model. With this as motivation, we train an instance of the full QG encoder--decoder model with the 
sentence encoder, but the sentence encoder is given the target question as input 
instead of the sentence. As expected, in this setting the model learns to generate 
questions very well because it "cheats" by taking the target as input. Next, the
trained sentence encoder --- that was trained by encoding the target questions ---
is decoupled from the full model for use in the following step.


In the second step, the question embeddings produced by the earlier trained encoder
are used as ground truth representations from which to maximize similarity. Now, a 
new sentence encoder is trained that takes as input a declarative sentence. The
new sentence encoder is trained to maximize the similarity between the input sentence's
embedding and the question embedding --- produced by the earlier trained encoder --- 
belonging to a specific question associated with the sentence. Specifically, loss
is calculated between a sentence embedding $s$ and a ground truth representation $q$
using a Cosine Embedding Loss defined by:
\begin{equation}
\text{loss}(s, q) = 1 - \text{cos}(s, q) 
\end{equation}
\noindent where cos$(s, q)$ is traditional cosine distance.

The pre-trained sentence encoder is then used to initialize the sentence encoder 
used in training a new instance of the full QG model.   


\begin{table}[h!t]
\begin{small}
\begin{tabular}
{@{} p{0.7in}|p{2.1in}@{}}
\toprule
\rowcolor[gray]{0.9} {\bf \# Model} & \multicolumn{1}{c}{ {\bf Sentences and Examples}}  \\ \hline

S3 & West got his big break in the year 2000, when he began to produce for artists on Roc-A-Fella Records. \\ \hline
Q10 Copy & when did west got his big break?   \\ 
Q11 No Copy & in what year did `` big break '' begin?   \\ \hline   \hline   
S4 & A high-definition remaster of the game, The Legend of Zelda: Twilight Princess HD, is being developed by Tantalus Media for the Wii U. \\ \hline
Q12 Copy &  who developed the legend of zelda?    \\ 
Q13 No Copy &  who is the creator of the soundtrack of mortal kombat?  \\ \hline   \hline   
S5 & Both six- and seven-track versions of the game's soundtrack were released on November 19, 2006, as part of a Nintendo Power promotion and bundled with replicas of the Master Sword and the Hylian Shield. \\ \hline
Q14  Copy & What was released on november 19, 2006?   \\ 
Q15 No Copy & What was released on september 17, 2006? \\      \hline \hline   
S6 & At the age of 10, West moved with his mother to Nanjing, China, where she was teaching at Nanjing University as part of an exchange program. \\ \hline
Q16 Copy & at what age did west move with his mother to nanjing?  \\
Q17 No Copy & at what age did von neumann teach at nanjing university? \\       \hline
\end{tabular}
\end{small}
\vspace{-.1in}
\caption{Question Generation with and without the Copy mechanism.\label{table:copy-examples}}
\vspace{-.1in}
\end{table}
\vspace{-.1in}

\section{Experimental Setup and Results}
\label{results-sec}

We conduct experiments exploring the effects of each of our model
enhancements and train and evaluate all the models using the SQuAD
dataset \cite{Rajpurkar_Zhang_Lopyrev_Liang_2016}.  We evaluate the
models using both automatic evaluation metrics and a human evaluation
using crowd sourced workers. In addition, we perform ablation tests to experiment with different feature settings. We name our two best models Focus and FocusCR, where FocusCR uses the coreference feature, and Focus does not.

\noindent \textbf{Dataset.}
SQuAD is a dataset of over one hundred thousand (document, question, answer) tuples. The documents are Wikipedia articles and the questions are created by crowd workers. Answers to the questions are subsequently created by a separate group of crowd workers who select as the question's answer a span of text from within the article. The creators of SQuAD keep part of the dataset private to be used as a hidden evaluation set in Question Answering tasks. For this work we use the roughly 92,000 examples that are publicly available. The 92,000 examples are partitioned into training (roughly 70k examples), development (roughly 10k examples), and test (roughly 11k examples) subsets. For the sake of comparison, we have used the same partitioning as \citet{Du_Shao_Cardie_2017} who have kindly made their data setting available on-line.

Using Stanford CoreNLP \cite{stanfordNER05,Manningetal14} 
the data is tokenized, and NER and Coreference Resolution 
are performed. All the feature used by our model are calculated
at this stage. Finally, the text is lowercased. We calculate
separate source and target vocabularies of size 45,000 and 28,000, 
respectfully. Tokens that fall out of vocabulary (OOV) are 
represented with a special \texttt{UNK} token. In retrospect, separate 
vocabularies are not necessary for this task. We remove examples 
that have sentences or questions over 100 and 50 words 
long, respectfully. 

\noindent \textbf{Model Implementation.}
Our model is implemented using PyTorch\footnote{\url{pytorch.org}} and OpenNMT-py\footnote{\url{github.com/OpenNMT/OpenNMT-py}} which is a PyTorch port of OpenNMT\cite{opennmt}. The encoder, decoder, and sentence encoder are multi-layer RNNs, each with two layers. We use bi-directional LSTM cells with 640 units. The model is trained using Dropout \cite{Srivastava_Hinton_Krizhevsky_Sutskever_Salakhutdinov_2014} of 0.3 between RNN layers. Word embeddings are initialized using Glove 300 dimensional word vectors \cite{Glove_Pennington_Socher_Manning_2014} that are not updated during training. The sentence encoder is initialized using the pre-training process described in Section~\ref{embed-sec}. All other model parameters are initialized using Glorot initialization \cite{Glorot_Bengio_2010}. 

The model parameters are optimized using Stochastic Gradient Descent with mini-batches of size 64. Beam search with five beams is used during inference and OOV words are replaced using the token of highest attention weight in the source sentence. We tune our model with the development dataset and select the model of lowest Perplexity to evaluate on the test dataset. 

\subsection{Automatic Evaluation}
\label{automatic-evaluation-sec}
We compare our system's results to that of several other QG systems. 
The rows of Table~\ref{table:multi-reference-results} with labels H\&S, Yuan, Du, and S\&X refer to the models presented in \citet{Heilman_Smith_2010,Yuan_Wang_Gulcehre_Sordoni_Bachman_Zhang_Subramanian_Trischler_2017,Du_Shao_Cardie_2017}, and \citet{Sachan_Xing_2018}, respectfully. Please refer to Section~\ref{rel-sec} Related Work for further details on each of these systems. The results of the H\&S system are reported in this work for the sake of comparison. The actual experiments were performed by \citet{Du_Shao_Cardie_2017} who describe the specific configuration of H\&S in greater detail.  

\noindent
\textbf{Results.} We use BLEU score \cite{papineni2002bleu} as an automatic evaluation
metric and compare directly to other work.  BLEU measures the
similarity between a generated text called a candidate and a set of
human written texts called a reference set. The score is calculated by
comparing the n-grams of the candidate with the n-grams of the
reference texts and then counting the number of matches.

Unfortunately there are inconsistencies in the method by which
previous works have used BLEU to evaluate QG models. Therefore, to accurately
compare BLEU scores, we evaluate our model using two different setups.
First, when calculating BLEU for a given hypothesis question, some
publications have used a reference set containing all the ground-truth
questions corresponding to the sentence from which the hypothesis was
generated. Table \ref{table:multi-reference-results} shows our model's results
compared to previous work using this setup of BLEU and the same
partitioning of the SQuAD dataset. 

Each of our models outperform 
previously published results in each of the BLEU, METEOR, and ROUGE
categories by a significant margin. FocusCR is the second highest performing system and achieves an impressive 
BLEU\_4 score of 19.86, which greatly improves on the third highest BLEU\_4
score of 14.37 belonging to S\&X. Focus gets a BLEU\_4 score of 19.98 and is the best performing system overall.

In the second setup, for a given hypothesis question,
\citet{Yuan_Wang_Gulcehre_Sordoni_Bachman_Zhang_Subramanian_Trischler_2017}
used a reference set containing only a single ground-truth question
that corresponds to the same sentence and answer span from which the
hypothesis was generated. We use this setup to evaluate 
our Focus and FocusCR models. The results are shown in Table 
\ref{table:single-reference-results}. Here, Focus and FocusCR are the same
models as shown in Table~\ref{table:multi-reference-results}, with 
the only difference being the setting under which they are evaluated. Again,
FocusCR achieves the second highest score and
Focus gets the highest BLEU\_4 score at 14.39.
While the datasets in aggregate are the same, our partitioning of training, 
development, and test datasets is different from that of
\citet{Yuan_Wang_Gulcehre_Sordoni_Bachman_Zhang_Subramanian_Trischler_2017}.

\begin{table}[t!]
\centering
\begin{tabular}{llll}
Model         				  & BLEU\_4         & METEOR         & ROUGE  \\
\hline
baseline      				  & 11.53          & 15.93           & 39.57 \\
H\&S\textsuperscript{*}       & 11.18          & 15.95          & 30.98 \\
Du  & 12.28          & 16.62          & 39.75 \\
S\&X  & 14.37          & 18.57          & 42.73 \\
\hline
FocusCR 			& 19.86 & 21.96 & \textbf{48.35} \\
\textbf{Focus} 		& \textbf{19.98} & \textbf{22.26} & 48.23
\end{tabular}
\caption{System performance in automatic evaluation.}
\label{table:multi-reference-results}
\end{table}

\begin{table}[]
\begin{center}
\begin{tabular}{ll}
Model    	  			 & BLEU\_4 \\ \hline
baseline 	  			 & 8.45   \\
Yuan     	  			 & 10.5   \\
FocusCR 				 & 14.16  \\
\textbf{Focus} 				& \textbf{14.39} \\
\end{tabular}
\caption{BLEU-4 scores when using answer-specific ground-truth questions as reference texts.}
\label{table:single-reference-results}
\end{center}
\end{table}

We perform ablation experiments to study the effects of each feature
incorporated into the model. The results of these experiments can 
be seen in Table~\ref{table:ablation-results}. With the exception of
the CoRef feature, each feature added produces an improvement in BLEU,
with the answer feature producing the greatest improvement. The 
copy-mechanism and sentence embedding, which is called Focus in the 
table, each increase performance further. 

We also examine the effect of pre-training the sentence encoder as described in Section~\ref{embed-sec}. In Table~\ref{table:ablation-results}, the Focus and FocusCR models use a pre-trained sentence encoder. The sentence encoder used by FocusCR-npt is not pre-trained. We find that the pre-training has a positive effect on BLEU scores with FocusCR-npt getting BLEU 13.99, compared to the FocusCR getting BLEU 14.16.

Table~\ref{table:ablation-results} suggests that the coreference
mechanism actually hurts performance as measured by BLEU but
the example shown in Table~\ref{table:coref-examples} and the
additional examples shown in Table~\ref{table:coref-examples-more} suggest
that it is very effective. 

Table~\ref{table:copy-examples} provides examples of the effect of the
copy mechanism.  Again, as with coreference, it is easy to see the
benefits of the copying mechanism qualititively.  For example,
in Q14 and Q15 the model can effectively copy the right date into
the question. In Q17, without
copying, the model finds an entity {\it von neumann} who is
associated with {\it teaching} and {\it university} in the model and
uses that entity to generate the question rather than {\it west}
the entity in context. 

\begin{table}[b]
\begin{center}
\begin{tabular}{lr}
Model       & Unique Q's \\ \hline
Baseline    & 6,595   \\
FocusCR     & 10,194   \\
Human 	    & 11,801  \\
\end{tabular}
\caption{Amount of unique questions generated.}
\label{table:question-quantity-results}
\end{center}
\end{table}

\noindent
\textbf{Question Diversity.} We are interested in how the new features effect the quantity of unique questions produced by our model. Therefore, we counted the number of unique questions output by the model when considering the entire testing set as inputs. Here, we measure similarity using a strict character match comparison. Table \ref{table:question-quantity-results} shows the results. We can see that the FocusCR model produces 10,194 unique questions, which is a 55\% increase over the 6,595 unique questions produced by the Baseline model. Although strict character matching is a crude method of measuring question similarity, we conclude that the features incorporated into the FocusCR model have a positive effect on the diversity of generated questions.


\begin{table}[]
\begin{tabular}{llll}
Model     	& BLEU\_4  & METEOR & ROUGE \\ \hline
baseline  	& 7.62  & 13.41 & 34.19 \\
+ Answer	& 11.15 & 16.64 & 40.39 \\
+ NER		& 11.54	& 16.94	& 40.93 \\
+ Case		& 11.56 & 16.98 & 40.96 \\
+ CoRef		& 10.28 & 16.14 & 39.22 \\
+ Copy		& 13.00	& 18.43 & 42.78 \\ 
\hline
FocusCR-npt     & 13.99 & - & - \\
FocusCR     & 14.16 & 19.24 & 43.07 \\
Focus     & 14.39 & 19.54 & 43.00 \\
\end{tabular}
\caption{Results of ablation test.
}
\label{table:ablation-results}
\vspace{-.1in}
\end{table}
\vspace{-.1in}
\subsection{Human Evaluation}
\label{human-results-sec}
We perform human evaluation using crowd workers on Amazon Mechanical Turk\footnote{\url{www.mturk.com}}. The Turkers rate a pool of questions constructed by randomly selecting questions and their associated text passages from the test set.  We select 114 questions each from the test dataset, the questions generated by the baseline model, and the questions generated by our FocusCR model. The questions are selected such that they all correspond to the same declarative sentence. In other words, we construct a set of 114 tuples where each tuple consists of one text passage, two model generated questions, and one human authored question. 

We use a qualification criteria to restrict the participation of Turkers in our evaluation study. The Turkers must have above 95\% HIT approval rate with at least 500 HITs previously approved. Furthermore, Turkers are required to be located in English speaking countries. Turkers recieved \$0.1 for completing each HIT. 

We closely follow the experiment design described by \citet{Heilman_Smith_2010_rating_computer_generated_questions}, who instruct Turkers to produce a single five-point quality rating per question. They provide Turkers with the following four reasons to downgrade a question: (Un)grammaticality, Incorrect Information, Vagueness, and Awkwardness. In our evaluation study, we use four categories of evaluation that resemble these criteria.  

Turkers are asked to rate each question across four categories: Grammaticality, Correct Information, Answerability, and Naturalness. Grammaticality encompasses things like adherence to rules of syntax, use of the wrong \textit{wh}-word, verb tense consistency, and overall legitimacy as an English sentence. The Correct Information category considers whether or not the question is related to the text passage (e.g., asking about Madonna when the passage is about Beyonce), implies something that is obviously incorrect, or contradicts information given in the text passage. The Answerability category reflects how much of the information required to correctly answer the question is contained within the text passage. Also, it considers whether or not the question has a clear answer, or is too vague (e.g., \textit{"What is it?"}). The Naturalness category reflects how natural the question reads and considers whether or not it has some awkward phrasing. The Naturalness category also encompasses any other problems in the question that do not fall in the previous categories.  


During evaluation, the Turker is presented with the text passage and its three corresponding questions in scrambled order. They are asked to give a rating from worst (1) to best (5) in each category for each question. Each HIT contains three text passages and a total of nine questions. Each HIT is assigned to three Turkers resulting in three ratings per question.


\noindent \textbf{Results.} Table~\ref{table:human-evaluation} shows an average of the ratings assigned by the Turkers in each category. Answerability is the category in which the FocusCR model has the greatest improvement over the Baseline. In this category, FocusCR receives an average rating of 4.13, compared to the baseline's average rating of 3.73. FocusCR also outperforms the Baseline model in the Correct Information category with average ratings 4.13 and 3.78, respectfully. In the Grammaticality and Naturalness categories the Baseline model has average ratings of 4.23 and 4.10, respectfully. The FocusCR model has average ratings of 4.20 and 4.09 in the Grammaticality and Naturalness categories. The human authored questions outperform both models by a significant margin in all categories. 

We note that there is only a slight difference between ratings achieved by the Baseline and FocusCR models in the Grammaticality and Naturalness categories. Yet, in both these categories the Baseline model slightly outperforms FocusCR. We suspect this is due to the brevity and generality of questions produced by the Baseline model. In contrast, FocusCR produces longer sentences with more information content and, at times, increasingly complex sentence structure. 

Next, we observe that the average rating of the human-authored questions are surprisingly low across all categories, but particularly in Naturalness with a rating of 4.36. We attribute this to the crowd-sourcing methodology used to create the original SQuAD dataset. Nevertheless, we hypothesize that the average ratings of the gold questions will increase with larger sample sizes in subsequent human evaluation studies. 

Inter-rater agreement was measured by comparing the Turkers' ratings to those of an expert annotator who is a native English speaking  graduate student in Computational Linguistics. The expert annotator rated a random sample of 60 questions using a private version of the HIT created on Mechanical Turk. Then, the arithmetic mean of the three Turker ratings was calculated for each question and category of evaluation. The Pearson correlation coefficient between the expert annotator's rating and the means of the Turker ratings was $r = 0.47$ for the Correct Information category, $r = 0.38$ for the Answerability category,  $r = 0.20$ for Grammaticality, and $r = 0.32$ for Naturalness. The significance of each correlation was calculated using a two-tailed test that resulted in $p < 0.01$ for each category. We observe a positive correlation between the expert annotator and the Turker ratings in each category, although some of the the correlation strengths are less than ideal, particularly in the Grammaticality category. The consistent positive correlation across each category and their statistical significance provide evidence  that the rating scheme is well defined, and that the Turkers are able to judge the quality of questions with relative reliability.


\begin{table}[h!t]
\begin{small}
\begin{center}
\begin{tabular}{lllll}

Model       & Grammar     & Info.   & Answer.   & Natural  \\ \hline
Baseline    & 4.23      & 3.78      & 3.73      & 4.10  \\
FocusCR     & 4.20      & 4.13      & 4.13      & 4.09  \\ \hline
Human       & \textbf{4.40}      & \textbf{4.40}     & \textbf{4.47}      & \textbf{4.36}  \\
\hline
\end{tabular}
\end{center}
\end{small}
\vspace{-.1in}
\caption{Human Evaluation Results \label{table:human-evaluation}}
\vspace{-.1in}
\end{table}

\begin{table}[h!t]
\begin{small}
\begin{tabular}
{@{} p{0.7in}|p{2.1in}@{}}
\toprule
\rowcolor[gray]{0.9} {\bf \# Model} & \multicolumn{1}{c}{ {\bf Sentences and Examples}}  \\ \hline
S7 & she publicly endorsed same sex marriage on march 26, 2013, after the supreme court debate on california 's proposition 8.\\ \hline
Q18 FocusCR & what did beyonce publicly support?  
  \\ 
Q19 Focus & what did madonna publicly endorsed on march 26, 2013?   \\ \hline   \hline   
S8 &  west is one of the best-selling artists of all time, having sold more than 32 million albums and 100 million digital downloads worldwide   \\ \hline  
Q20 FocusCR &how many grammy awards did he win?  \\
Q21 Focus & how many grammy awards has madonna won? \\   \hline   
\end{tabular}
\end{small}
\vspace{-.1in}
\caption{Additional Coreference Examples\label{table:coref-examples-more}}
\end{table}
\vspace{-.1in}

\section{Related Work}
\label{rel-sec}


Much of the work on automatic question generation has been motivated by helping teachers in test creation \cite{Mitkov_Ha_2003, Heilman_Smith_2010, Labutov_Basu_Vanderwende_2015, Araki_Rajagopal_Sankaranarayanan_Holm_Yamakawa_Mitamura_2016, Chinkina_Meurers_2017}.
Questions play an essential role in knowledge acquisition and assessment.
It is standard practice for teachers to assess students' reading comprehension through question answering. 
Automatic question generation has the potential to assist teachers in the test creation process, thereby freeing teachers to spend more time on other aspects of the education process, and reducing the cost of receiving an education.   

Automatic question generation has the potential to be useful in the areas of automatic Question Answering (QA) and Machine Comprehension of text. Recently, large datasets such as SQuAD \cite{Rajpurkar_Zhang_Lopyrev_Liang_2016}, and MS MARCO \cite{Nguyen_Rosenberg_Song_Gao_Tiwary_Majumder_Deng_2016} have facilitated advances in both areas. 
These datasets are expensive to create and consist of human authored (document, question, answer) triples with questions and answers either being collected from the web or created by crowd workers. Automatic Question Generation methods can be used to cheaply supplement resources available to QA models, further assisting in advancing QA capabilities. Indeed, \citet{Sachan_Xing_2018} have recently shown that a joint QA-QG model is able to achieve state-of-the art results on a variety of different QA related tasks.

Sequence-to-sequence Neural Network models have been shown to be effective at a
variety of other NLP problems
\cite{Bahdanau_Cho_Bengio_2014,rush2015neural,juraska2018deep}, and
recent work has also applied them to QG \cite{Du_Shao_Cardie_2017,
  Zhou_Yang_Wei_Tan_Bao_Zhou_2017,
  Yuan_Wang_Gulcehre_Sordoni_Bachman_Zhang_Subramanian_Trischler_2017}.
As in other recent work on QG, we use an attentional Recurrent Neural
Network encoder--decoder model that is similar to the model of
\citet{Bahdanau_Cho_Bengio_2014}.  In this approach, the QG task is
cast as a sequence-to-sequence language modeling task. The input
sentence, represented as a series of words, is mapped to an output
series of words representing a question.  Sequence-to-sequence models
have several advantages over previous rule-based approaches to
QG. First, they eliminate the need for large hand-crafted rule sets
-- the model automatically learns how to perform the subtasks of
Question Representation Construction and Question Realization. Another
advantage is that the model does not rely on domain-specific
keywords. In fact, in this approach the model is trained on examples
from a variety of topics and then evaluated on examples from
previously unseen topic domains.

Among the numerous approaches to question generation, our
work is most similar to recent work applying neural network models to
the task of generating short answer factoid questions.

\citet{Yuan_Wang_Gulcehre_Sordoni_Bachman_Zhang_Subramanian_Trischler_2017} developed a Recurrent Neural Network (RNN) sequence-to-sequence model that generates questions from an input sentence. Their model is trained using supervised learning combined with reinforcement learning to maximize several auxiliary goals, including performance of a QA model on generated questions. 

\citet{Du_Shao_Cardie_2017}
present an attentional sequence-to-sequence model for question generation. Their model is similar to our baseline model but with one key difference: their model uses paragraph-level information in addition to sentence-level information. They use an RNN encoder to embed the paragraph surrounding the sentence that contains the answer. Then the decoder's hidden state is initialized with the concatenation of the encoder's outputs and the paragraph embedding. 

\citet{Sachan_Xing_2018} present an ensemble model that jointly learns both QA and QG tasks. The QG model is an RNN sequence-to-sequence model similar to that proposed by \citet{Du_Shao_Cardie_2017}. First the QA and QG models are trained independent of each other on the labeled corpus. Then the QG model is used to create more questions from unlabeled data that are then answered by the QA model. A question selection oracle selects --- based on several heuristics --- a subsample of questions upon which to stochastically update each model. This process is repeated until both models cease to show improvement.  

\citet{Heilman_Smith_2010} present a system that generates fact-based questions similar to those in SQuAD using an "overgenerate-and-rank" strategy. Their system generates questions through use of hand crafted rules that operate on declarative sentences, transforming them into questions. In order to control quality, the output questions are filtered through a logistic regression model that ranks the questions on acceptability. 

\section{Conclusion and Future Work}
\label{conc-sec}
We propose an encoder--decoder model for automatic 
generation of factual questions.  We create a novel Neural Network architecture that uses two source sequence encoders; the first encoder being at the token level, and the second being at the sentence level. This enables the decoder to take into account word meaning and sentence meaning information while making decoding decisions. Also, the encoders are able to produce diverse encodings based on an answer focus feature.
We demonstrate that this new model greatly improves on the state of the art in Question Generation when evaluated using automatic methods. We show that incorporating linguistic features into our model improves question generation performance as well. Lastly, a human evaluation confirms the improvement in quality of generated questions.
 
 Currently, our system generates only factual questions for expository text. In future work we plan to explore question generation on other categories of text such as narrative discourse. One limitation of our system is that it relies on the existence of previously created answer phrases. Therefore, we would like to investigate methods of automatically extracting answer candidates from text, thus facilitating QG experiments on other categories of text that do not currently have large question-answer datasets.    
 
\section{Acknowledgements}
This work was supported by NSF Cyberlearning EAGER grant IIS-1748056, NSF Robust Intelligence grant IIS-1302668-002, and an Amazon Alexa Prize 2017 Gift and 2018 Grant awarded to the Natural Language and Dialogue Systems Lab at UC Santa Cruz. We thank Staunton Sample for his careful proofreading.

\bibliography{inlg18_qg}

\begin{thebibliography}{34}
\expandafter\ifx\csname natexlab\endcsname\relax\def\natexlab#1{#1}\fi

\bibitem[{Agarwal et~al.(2011)Agarwal, Shah, and
  Mannem}]{Agarwal_Shah_Mannem_2011}
Manish Agarwal, Rakshit Shah, and Prashanth Mannem. 2011.
\newblock Automatic question generation using discourse cues.
\newblock In \emph{Proceedings of the 6th Workshop on Innovative Use of NLP for
  Building Educational Applications}, page 1–9. Association for Computational
  Linguistics.

\bibitem[{Araki et~al.(2016)Araki, Rajagopal, Sankaranarayanan, Holm, Yamakawa,
  and Mitamura}]{Araki_Rajagopal_Sankaranarayanan_Holm_Yamakawa_Mitamura_2016}
Jun Araki, Dheeraj Rajagopal, Sreecharan Sankaranarayanan, Susan Holm, Yukari
  Yamakawa, and Teruko Mitamura. 2016.
\newblock Generating questions and multiple-choice answers using semantic
  analysis of texts.
\newblock In \emph{Proceedings of COLING 2016, the 26th International
  Conference on Computational Linguistics: Technical Papers}, page 1125–1136.

\bibitem[{Bahdanau et~al.(2014)Bahdanau, Cho, and
  Bengio}]{Bahdanau_Cho_Bengio_2014}
Dzmitry Bahdanau, Kyunghyun Cho, and Yoshua Bengio. 2014.
\newblock \href {http://arxiv.org/abs/1409.0473} {Neural machine translation by
  jointly learning to align and translate}.
\newblock \emph{arXiv:1409.0473 [cs, stat]}.
\newblock ArXiv: 1409.0473.

\bibitem[{Banerjee and Lavie(2005)}]{banerjee2005meteor}
Satanjeev Banerjee and Alon Lavie. 2005.
\newblock Meteor: An automatic metric for mt evaluation with improved
  correlation with human judgments.
\newblock In \emph{Proceedings of the acl workshop on intrinsic and extrinsic
  evaluation measures for machine translation and/or summarization}, pages
  65--72.

\bibitem[{Chali and Hasan(2015)}]{Chali_Hasan_2015}
Yllias Chali and Sadid~A. Hasan. 2015.
\newblock \href {https://doi.org/10.1162/COLI_a_00206} {Towards
  topic-to-question generation}.
\newblock \emph{Computational Linguistics}, 41(1):1–20.

\bibitem[{Chinkina and Meurers(2017)}]{Chinkina_Meurers_2017}
Maria Chinkina and Detmar Meurers. 2017.
\newblock Question generation for language learning: From ensuring texts are
  read to supporting learning.
\newblock In \emph{Proceedings of the 12th Workshop on Innovative Use of NLP
  for Building Educational Applications}, page 334–344.

\bibitem[{Du et~al.(2017)Du, Shao, and Cardie}]{Du_Shao_Cardie_2017}
Xinya Du, Junru Shao, and Claire Cardie. 2017.
\newblock \href {http://arxiv.org/abs/1705.00106} {Learning to ask: Neural
  question generation for reading comprehension}.
\newblock \emph{arXiv:1705.00106 [cs]}.
\newblock ArXiv: 1705.00106.

\bibitem[{Finkel et~al.(2005)Finkel, Grenager, and Manning}]{stanfordNER05}
Jenny~Rose Finkel, Trond Grenager, and Christopher~D. Manning. 2005.
\newblock Incorporating non-local information into information extraction
  systems by gibbs sampling.
\newblock In \emph{Proc. of the 43nd Annual Meeting of the Association for
  Computational Linguistics (ACL 2005)}, pages 363--370.

\bibitem[{Glorot and Bengio(2010)}]{Glorot_Bengio_2010}
Xavier Glorot and Yoshua Bengio. 2010.
\newblock Understanding the difficulty of training deep feedforward neural
  networks.
\newblock In \emph{Proceedings of the thirteenth international conference on
  artificial intelligence and statistics}, page 249–256.

\bibitem[{Heilman and Smith(2010{\natexlab{a}})}]{Heilman_Smith_2010}
Michael Heilman and Noah~A. Smith. 2010{\natexlab{a}}.
\newblock \href {http://dl.acm.org/citation.cfm?id=1857999.1858085} {Good
  question! statistical ranking for question generation}.
\newblock In \emph{Human Language Technologies: The 2010 Annual Conference of
  the North American Chapter of the Association for Computational Linguistics},
  HLT ’10, page 609–617. Association for Computational Linguistics.

\bibitem[{Heilman and
  Smith(2010{\natexlab{b}})}]{Heilman_Smith_2010_rating_computer_generated_questions}
Michael Heilman and Noah~A. Smith. 2010{\natexlab{b}}.
\newblock \href {http://dl.acm.org/citation.cfm?id=1866696.1866701} {Rating
  computer-generated questions with mechanical turk}.
\newblock In \emph{Proceedings of the NAACL HLT 2010 Workshop on Creating
  Speech and Language Data with Amazon’s Mechanical Turk}, CSLDAMT ’10,
  page 35–40. Association for Computational Linguistics.

\bibitem[{Hochreiter and Schmidhuber(1997)}]{hochreiter1997long}
Sepp Hochreiter and J{\"u}rgen Schmidhuber. 1997.
\newblock Long short-term memory.
\newblock \emph{Neural computation}, 9(8):1735--1780.

\bibitem[{Juraska et~al.(2018)Juraska, Karagiannis, Bowden, and
  Walker}]{juraska2018deep}
Juraj Juraska, Panagiotis Karagiannis, Kevin Bowden, and Marilyn Walker. 2018.
\newblock A deep ensemble model with slot alignment for sequence-to-sequence
  natural language generation.
\newblock In \emph{Proceedings of the 2018 Conference of the North American
  Chapter of the Association for Computational Linguistics: Human Language
  Technologies, Volume 1 (Long Papers)}, volume~1, pages 152--162.

\bibitem[{Klein et~al.(2017)Klein, Kim, Deng, Senellart, and Rush}]{opennmt}
Guillaume Klein, Yoon Kim, Yuntian Deng, Jean Senellart, and Alexander~M. Rush.
  2017.
\newblock \href {https://doi.org/10.18653/v1/P17-4012} {Open{NMT}: Open-source
  toolkit for neural machine translation}.
\newblock In \emph{Proc. ACL}.

\bibitem[{Labutov et~al.(2015)Labutov, Basu, and
  Vanderwende}]{Labutov_Basu_Vanderwende_2015}
Igor Labutov, Sumit Basu, and Lucy Vanderwende. 2015.
\newblock Deep questions without deep understanding.
\newblock In \emph{Proceedings of the 53rd Annual Meeting of the Association
  for Computational Linguistics and the 7th International Joint Conference on
  Natural Language Processing (Volume 1: Long Papers)}, volume~1, page
  889–898.

\bibitem[{Lin(2004)}]{lin2004rouge}
Chin-Yew Lin. 2004.
\newblock Rouge: A package for automatic evaluation of summaries.
\newblock \emph{Text Summarization Branches Out}.

\bibitem[{Luong et~al.(2015)Luong, Pham, and Manning}]{luong2015effective}
Minh-Thang Luong, Hieu Pham, and Christopher~D Manning. 2015.
\newblock Effective approaches to attention-based neural machine translation.
\newblock \emph{arXiv preprint arXiv:1508.04025}.

\bibitem[{Manning et~al.(2014)Manning, Surdeanu, Bauer, Finkel, Bethard, and
  McClosky}]{Manningetal14}
Christopher~D Manning, Mihai Surdeanu, John Bauer, Jenny~Rose Finkel, Steven
  Bethard, and David McClosky. 2014.
\newblock The {Stanford} {CoreNLP} natural language processing toolkit.
\newblock In \emph{ACL (System Demonstrations)}, pages 55--60.

\bibitem[{Mitkov and Ha(2003)}]{Mitkov_Ha_2003}
Ruslan Mitkov and Le~An Ha. 2003.
\newblock \href {https://doi.org/10.3115/1118894.1118897} {Computer-aided
  generation of multiple-choice tests}.
\newblock In \emph{Proceedings of the HLT-NAACL 03 Workshop on Building
  Educational Applications Using Natural Language Processing - Volume 2},
  HLT-NAACL-EDUC ’03, page 17–22. Association for Computational
  Linguistics.

\bibitem[{Mostow and Chen(2009)}]{MOSTOW_CHEN_2009}
Jack Mostow and Wei Chen. 2009.
\newblock Generating instruction automatically for the reading strategy of
  self-questioning.
\newblock In \emph{The 14th International Conference on Artificial Intelligence
  in Education}, page~8.

\bibitem[{Nallapati et~al.(2016)Nallapati, Zhou, santos, Gulcehre, and
  Xiang}]{Nallapati_Zhou_santos_Gulcehre_Xiang_2016}
Ramesh Nallapati, Bowen Zhou, Cicero Nogueira~dos santos, Caglar Gulcehre, and
  Bing Xiang. 2016.
\newblock \href {http://arxiv.org/abs/1602.06023} {Abstractive text
  summarization using sequence-to-sequence rnns and beyond}.
\newblock \emph{arXiv:1602.06023 [cs]}.
\newblock ArXiv: 1602.06023.

\bibitem[{Nguyen et~al.(2016)Nguyen, Rosenberg, Song, Gao, Tiwary, Majumder,
  and Deng}]{Nguyen_Rosenberg_Song_Gao_Tiwary_Majumder_Deng_2016}
Tri Nguyen, Mir Rosenberg, Xia Song, Jianfeng Gao, Saurabh Tiwary, Rangan
  Majumder, and Li~Deng. 2016.
\newblock \href {http://arxiv.org/abs/1611.09268} {Ms marco: A human generated
  machine reading comprehension dataset}.
\newblock \emph{arXiv:1611.09268 [cs]}.
\newblock ArXiv: 1611.09268.

\bibitem[{Olney et~al.(2012)Olney, Graesser, and
  Person}]{Olney_Graesser_Person_2012}
Andrew~M. Olney, Arthur~C. Graesser, and Natalie~K. Person. 2012.
\newblock Question generation from concept maps.
\newblock \emph{Dialogue \& Discourse}, 3(2):75–99.

\bibitem[{Papineni et~al.(2002)Papineni, Roukos, Ward, and
  Zhu}]{papineni2002bleu}
Kishore Papineni, Salim Roukos, Todd Ward, and Wei-Jing Zhu. 2002.
\newblock Bleu: a method for automatic evaluation of machine translation.
\newblock In \emph{Proceedings of the 40th annual meeting on association for
  computational linguistics}, pages 311--318. Association for Computational
  Linguistics.

\bibitem[{Pennington et~al.(2014)Pennington, Socher, and
  Manning}]{Glove_Pennington_Socher_Manning_2014}
Jeffrey Pennington, Richard Socher, and Christopher Manning. 2014.
\newblock Glove: Global vectors for word representation.
\newblock In \emph{Proceedings of the 2014 conference on empirical methods in
  natural language processing (EMNLP)}, page 1532–1543.

\bibitem[{Rajpurkar et~al.(2016)Rajpurkar, Zhang, Lopyrev, and
  Liang}]{Rajpurkar_Zhang_Lopyrev_Liang_2016}
Pranav Rajpurkar, Jian Zhang, Konstantin Lopyrev, and Percy Liang. 2016.
\newblock \href {http://arxiv.org/abs/1606.05250} {Squad: 100,000+ questions
  for machine comprehension of text}.
\newblock \emph{arXiv:1606.05250 [cs]}.
\newblock ArXiv: 1606.05250.

\bibitem[{Rus et~al.(2010)Rus, Wyse, Piwek, Lintean, Stoyanchev, and
  Moldovan}]{Rus_Wyse_Piwek_Lintean_Stoyanchev_Moldovan_2010}
Vasile Rus, Brendan Wyse, Paul Piwek, Mihai Lintean, Svetlana Stoyanchev, and
  Cristian Moldovan. 2010.
\newblock \emph{The first question generation shared task evaluation
  challenge}. Association for Computational Linguistics.

\bibitem[{Rush et~al.(2015)Rush, Chopra, and Weston}]{rush2015neural}
Alexander~M Rush, Sumit Chopra, and Jason Weston. 2015.
\newblock A neural attention model for abstractive sentence summarization.
\newblock \emph{arXiv preprint arXiv:1509.00685}.

\bibitem[{Sachan and Xing(2018)}]{Sachan_Xing_2018}
Mrinmaya Sachan and Eric Xing. 2018.
\newblock \href {https://doi.org/10.18653/v1/N18-1058} {Self-training for
  jointly learning to ask and answer questions}.
\newblock \emph{Proceedings of the 2018 Conference of the North American
  Chapter of the Association for Computational Linguistics: Human Language
  Technologies, Volume 1 (Long Papers)}, 1:629–640.

\bibitem[{See et~al.(2017)See, Liu, and Manning}]{see2017get}
Abigail See, Peter~J Liu, and Christopher~D Manning. 2017.
\newblock Get to the point: Summarization with pointer-generator networks.
\newblock In \emph{Proceedings of the 55th Annual Meeting of the Association
  for Computational Linguistics (Volume 1: Long Papers)}, volume~1, pages
  1073--1083.

\bibitem[{Srivastava et~al.(2014)Srivastava, Hinton, Krizhevsky, Sutskever, and
  Salakhutdinov}]{Srivastava_Hinton_Krizhevsky_Sutskever_Salakhutdinov_2014}
Nitish Srivastava, Geoffrey Hinton, Alex Krizhevsky, Ilya Sutskever, and Ruslan
  Salakhutdinov. 2014.
\newblock Dropout: a simple way to prevent neural networks from overfitting.
\newblock \emph{The Journal of Machine Learning Research}, 15(1):1929–1958.

\bibitem[{Sutskever et~al.(2014)Sutskever, Vinyals, and
  Le}]{sutskever2014sequence}
Ilya Sutskever, Oriol Vinyals, and Quoc~V Le. 2014.
\newblock Sequence to sequence learning with neural networks.
\newblock In \emph{Advances in neural information processing systems}, pages
  3104--3112.

\bibitem[{Yuan et~al.(2017)Yuan, Wang, Gulcehre, Sordoni, Bachman, Zhang,
  Subramanian, and
  Trischler}]{Yuan_Wang_Gulcehre_Sordoni_Bachman_Zhang_Subramanian_Trischler_2017}
Xingdi Yuan, Tong Wang, Caglar Gulcehre, Alessandro Sordoni, Philip Bachman,
  Saizheng Zhang, Sandeep Subramanian, and Adam Trischler. 2017.
\newblock Machine comprehension by text-to-text neural question generation.
\newblock In \emph{Proceedings of the 2nd Workshop on Representation Learning
  for NLP}, page 15–25.

\bibitem[{Zhou et~al.(2017)Zhou, Yang, Wei, Tan, Bao, and
  Zhou}]{Zhou_Yang_Wei_Tan_Bao_Zhou_2017}
Qingyu Zhou, Nan Yang, Furu Wei, Chuanqi Tan, Hangbo Bao, and Ming Zhou. 2017.
\newblock \href {http://arxiv.org/abs/1704.01792} {Neural question generation
  from text: A preliminary study}.
\newblock \emph{arXiv:1704.01792 [cs]}.
\newblock ArXiv: 1704.01792.

\end{thebibliography}
\bibliographystyle{acl_natbib}

\end{document}